\documentclass[11pt]{article}

\usepackage[margin=1in]{geometry}
\usepackage[utf8]{inputenc}
\usepackage[T1]{fontenc}
\usepackage{amsmath,amssymb,amsthm}
\usepackage{booktabs}
\usepackage{graphicx}
\usepackage{array}
\usepackage{enumitem}
\usepackage{xcolor}
\usepackage{hyperref}
\usepackage{microtype}
\usepackage{caption}
\usepackage{subcaption}
\usepackage{longtable}
\usepackage{placeins}
\usepackage{float}

\hypersetup{
  colorlinks=true,
  linkcolor=blue,
  citecolor=blue,
  urlcolor=blue
}

\newcommand{\norm}[1]{\left\lVert #1 \right\rVert}
\newcommand{\argmax}{\operatorname*{arg\,max}}
\newcommand{\argmin}{\operatorname*{arg\,min}}

\newcommand{\sparsemax}{\operatorname{sparsemax}}
\newcommand{\BandNorm}{\operatorname{BandNorm}}

\title{Towards Verifiable Transformers:\\Solver-Checkable Circuit Explanations}

\author{Neel Somani\\
Independent Researcher\\
\texttt{neeljaysomani@gmail.com}}

\date{July 28, 2026}

\begin{document}

\maketitle

\begin{abstract}
Mechanistic interpretability typically discovers circuits and then argues what they do from examples and ablations. We introduce \emph{Verifiable Transformers}, a framework for turning task-localized circuits into bounded, solver-checkable claims: projected functional equivalence, task-relevant invariance, edge necessity, and robustness to continuous final-residual perturbations. At small scale, we directly verify all four properties for quote-closing and bracket-type circuits, including program-mediated circuits whose attention selection is entirely symbolic.

At GPT-2 scale, we remove LayerNorm from a sparsemax/LeakyReLU model after training with a $+0.0087$ OpenWebText loss increase, replace retained attention heads with synthesized restricted-DSL programs, and calibrate only program-local readouts while freezing and hashing every other parameter. The resulting three-edge quote circuit (embedding $\rightarrow$ MLP 0 $\rightarrow$ program head $\rightarrow$ logits) verifies all four properties over a hash-pinned 1,280-prompt domain in linear real arithmetic: 1,280/1,280 equivalence and invariance, 640 edge-necessity witnesses per edge, and robustness at $\epsilon=0.01$ with minimum certified radius $0.01515$.

For two alphabet variants of the same opener-detection/copy-type task, untouched gates expose a localization frontier: bracket-type extraction is exact only with all 144 heads, while the constructed quote artifact verifies with three edges. Naive discovery-based verification failed for measured reasons; at scale, we find we must build the object we can verify. The verified object is a calibrated artifact, not the unmodified model, and all claims are bounded to declared domains.
\end{abstract}

\section{Introduction}

Mechanistic interpretability often treats trained Transformers as objects to be reverse engineered after training. This empirical program has produced localized attention heads, sparse features, and hand-analyzed circuits for algorithmic behavior \cite{olsson2022induction,nanda2023progress,gao2025weightsparse}. But Transformer models are engineered artifacts. Besides improving post-hoc analysis, we can design and modify models so that specific mechanistic claims become easier to state, encode, and check.

This paper studies that goal under the name \emph{Verifiable Transformers}. The target is not end-to-end verification of arbitrary language models. It is a narrower class of circuit-level claims:

\begin{itemize}[leftmargin=*]
    \item Does an extracted circuit make the same projected task decision as a symbolic reference program throughout a declared domain?
    \item Is each retained edge behaviorally necessary under a specified ablation semantics?
    \item Is the circuit invariant to task-irrelevant input changes?
    \item Is the projected decision stable under every perturbation in a continuous $\ell_\infty$ ball at a specified internal interface?
    \item Does a compact symbolic explanation exactly describe the circuit, or can a solver return a counterexample?
\end{itemize}

We study both direct and constructive paths to these guarantees. Small SMT-representable circuits are encoded directly. At GPT-2 scale, we remove normalization after training, replace selected neural attention heads with synthesized programs, constrain calibration so that it cannot create a new non-program bypass, and verify a three-edge quote-closing circuit over a declared 1,280-prompt domain. The central methodological lesson is constructive: after discovery-based routes fail for measured reasons, we build the object we can verify.

\begin{figure}[H]
    \centering
    \includegraphics[width=0.98\linewidth]{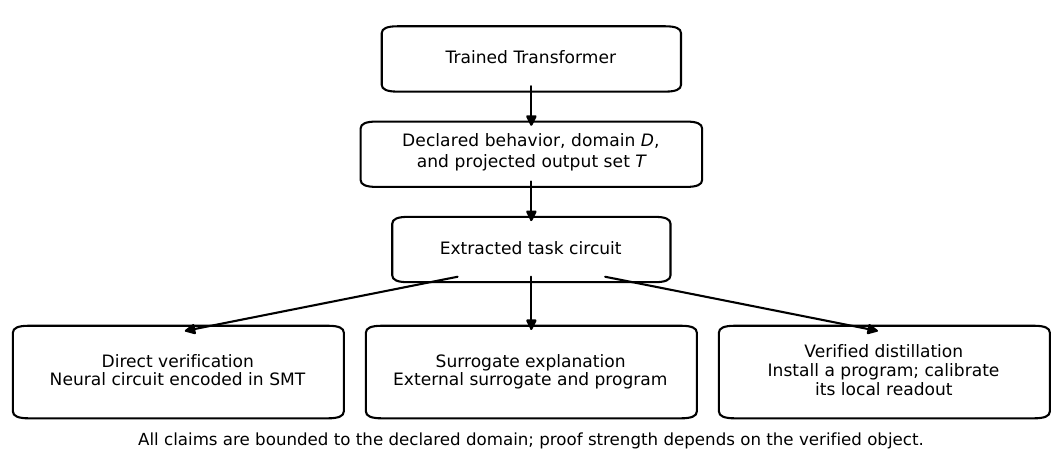}
    \caption{Three settings for circuit-level verification. Direct verification encodes a neural circuit. Surrogate-mediated verification checks an external abstraction. Verified distillation replaces the relevant mechanism with a symbolic program and constrains calibration so that the program remains causally load-bearing.}
    \label{fig:modes}
\end{figure}

Figure~\ref{fig:modes} distinguishes the three verification settings used in the paper. Direct verification encodes a neural circuit itself. Surrogate-mediated verification checks an external symbolic explanation of a hard-to-encode circuit. Verified distillation changes the artifact: a symbolic program is installed as the circuit's attention mechanism, and only parameters whose contribution vanishes under program lesion are calibrated.

\subsection{Contributions}

The paper makes six main contributions.

\begin{enumerate}[leftmargin=*]
    \item \textbf{Circuit-level verification framework.} We formalize projected functional equivalence, task-relevant invariance, edge necessity, and final-residual robustness for extracted Transformer circuits over declared bounded domains.
    \item \textbf{Small-scale direct and program-mediated verification.} On quote-closing and bracket-type tasks, small circuits verify all four properties. Replacing neural attention with restricted programs removes the active bilinear QK terms and further reduces encoding cost.
    \item \textbf{Normalization removal for a sparsemax GPT-2 variant.} We show that all 25 LayerNorm instances can be attenuated and folded out of a sparsemax/LeakyReLU GPT-2-scale model for only $+0.0087$ OpenWebText loss relative to its LayerNorm source. The norm-free model is also strictly easier to prove robust because branch-certificate unknowns disappear.
    \item \textbf{Localization frontier.} Across three preregistered domain generations with untouched gates, sparse circuits are repeatedly near-exact but not exact beyond their extraction data. An exact bracket-type circuit ultimately retains all 144 attention heads, while quote closing can be constructively localized to three edges. The tasks implement the same high-level opener-detection/copy-type algorithm over different alphabets.
    \item \textbf{Causally constrained calibration.} We synthesize program heads, freeze and hash every non-program parameter, and train only program-local readouts. Under program lesion, every trainable contribution vanishes, so calibration cannot create a new bypass; any surviving route is a measured property of the frozen base model.
    \item \textbf{GPT-2-scale verified circuit.} The final quote circuit verifies all four properties over a hash-pinned 1,280-prompt domain. A constant-folded exact-rational verifier reduces the battery to ground linear-real-arithmetic checks, completing the full battery in 66.61 seconds.
\end{enumerate}

\subsection{Claim discipline}

The strongest result is about a \emph{constructed and calibrated artifact}, not post-hoc verification of the unmodified GPT-2-scale model. The domain is part of the claim. The 1,280 prompts are unique, frozen, and hash-pinned; no arbitrary-length or out-of-domain correctness is claimed. A preserved solver counterexample outside this domain marks the boundary constructively.

The paper also does not claim full-vocabulary equivalence, global circuit minimality, or absence of native redundant mechanisms. Edge necessity is relative to zero ablation and the retained circuit. The constrained calibration procedure prevents training from creating a new bypass, but a pre-existing frozen route remains measurable and is reported explicitly.

\section{Formal Framework}

\subsection{Projected decisions and extracted circuits}

Let $M$ be a trained model, $D$ a declared finite input domain, and $T$ a task-specific candidate-token set. For a model or circuit $F$, let $F_T(x)$ be the candidate logits and define
\begin{equation}
    d_T(F,x) := \argmax_{t\in T} F_t(x).
    \label{eq:projected-decision}
\end{equation}
Let $P:D\rightarrow T$ be a symbolic reference program. The output projection is part of the specification: quote closing asks only whether the next token is a single or double quote, and bracket type asks only whether it is a square or curly closing delimiter.

Let $C_E$ be a circuit obtained by retaining edge set $E$ in a coarse or per-head computational graph and setting deleted edge contributions to zero. Retained nodes use the model's trained parameters. For $e\in E$, $C_{E\setminus\{e\}}$ denotes the same circuit with edge $e$ deleted. If extraction is intended to preserve the full model, its projected faithfulness condition is
\begin{equation}
    \forall x\in D,\qquad d_T(C_E,x)=d_T(M,x).
    \label{eq:circuit-faithfulness}
\end{equation}
A circuit optimized only against $P$ is a task circuit; it need not be the unmodified model's native mechanism.

\subsection{Four verification properties}

Table~\ref{tab:properties} gives the four properties used throughout the paper. In the robustness statement, $r_E(x)$ is the final residual before the final normalization, when one exists, and $G_T(r)$ is the projected decision after the final normalization and unembedding.

\begin{table}[H]
\centering
\small
\resizebox{\linewidth}{!}{%
\begin{tabular}{>{\raggedright\arraybackslash}p{0.19\linewidth}>{\raggedright\arraybackslash}p{0.36\linewidth}>{\raggedright\arraybackslash}p{0.39\linewidth}}
\toprule
Property & Meaning & Formal statement \\
\midrule
Projected functional equivalence & The circuit agrees with the symbolic reference program throughout the declared domain. & $\forall x\in D,\ d_T(C_E,x)=P(x)$ \\
Edge necessity & Every retained edge is load-bearing for at least one input under zero ablation. This is not global minimality. & $\forall e\in E,\ \exists x\in D:\ d_T(C_E,x)\neq d_T(C_{E\setminus\{e\}},x)$ \\
Task-relevant invariance & Inputs related by a declared task-irrelevant relation $R$ receive the same projected decision. & $\forall x,x'\in D,\ R(x,x')\Rightarrow d_T(C_E,x)=d_T(C_E,x')$ \\
Final-residual robustness & Every perturbation inside an $\ell_\infty$ ball at the final residual preserves the projected decision. & $\forall x\in D,\ \forall\eta,\ \norm{\eta}_\infty\leq\epsilon\Rightarrow G_T(r_E(x)+\eta)=G_T(r_E(x))$ \\
\bottomrule
\end{tabular}%
}
\caption{Circuit-level properties. The domain, candidate-token projection, relation $R$, perturbation interface, and ablation semantics are all part of the claim.}
\label{tab:properties}
\end{table}

\subsection{What is verified}

\paragraph{Direct verification.}
When $C_E$ is tractably SMT-representable, the verifier encodes the circuit and the negation of a target property. For functional equivalence it asks whether
\begin{equation}
    \exists x\in D:\ d_T(C_E,x)\neq P(x).
    \label{eq:direct-query}
\end{equation}
Unsatisfiability proves the property over $D$; satisfiability returns a counterexample. Proofs are with respect to the exported exact real or rational circuit, not bit-for-bit PyTorch floating-point execution. Exhaustive PyTorch checks and encoder-agreement tests guard the export boundary.

\paragraph{Surrogate-mediated verification.}
For a hard-to-encode circuit, an SMT-encodable surrogate $S$ is checked against a finite teacher relation, and a symbolic program $P$ is checked against $S$:
\begin{align}
    \forall x\in D,&\quad d_T(S,x)=d_T(C_E,x), \label{eq:surrogate-fidelity}\\
    \forall x\in D,&\quad P(x)=d_T(S,x). \label{eq:program-equivalence}
\end{align}
When both links are exhaustive on the same domain, $P$ captures the circuit's projected behavior there. The proof object remains the surrogate or finite teacher relation rather than an exact encoding of the original neural computation.

\paragraph{Verified distillation.}
Verified distillation installs a restricted symbolic attention program $H$ inside a new artifact $\widetilde M$. Let $\theta_H$ denote program-local trainable parameters and $\theta_{\neg H}$ all other parameters. Calibration freezes and hashes $\theta_{\neg H}$ and permits training only in a set whose contribution vanishes when the program is lesioned. If $L_H$ zeros all program-head contributions, then by construction
\begin{equation}
L_H\!\left(\widetilde M_{\theta_{\neg H},\theta_H}\right)(x)
=
L_H\!\left(\widetilde M_{\theta_{\neg H},\theta_H^{(0)}}\right)(x)
\quad\text{for every }x,
\label{eq:lesion-identity}
\end{equation}
where $\theta_H^{(0)}$ is the zero-step program installation. Equation~\eqref{eq:lesion-identity} is an identity, not a regularizer: calibration cannot create a new non-program bypass. It does not imply that the frozen base model lacks a native redundant route; that route is measured separately.

\subsection{Norm-free robustness}

BandNorm is continuous and piecewise-affine, but a branch-certified verifier may return unknown if a perturbation ball crosses a branch boundary. Removing the final normalization yields a simpler specialization. Let the candidate logit for token $t$ be
\begin{equation}
    \ell_t(r)=u_t^\top r+b_t.
\end{equation}
If $y=d_T(C_E,x)$, its margin over competitor $t$ is
\begin{equation}
    m_{y,t}(x)=\ell_y(r_E(x))-\ell_t(r_E(x)).
\end{equation}
For $\norm{\eta}_\infty\leq\epsilon$, the worst-case margin reduction is $\epsilon\norm{u_y-u_t}_1$. Therefore the decision is robust exactly when
\begin{equation}
    m_{y,t}(x)>\epsilon\norm{u_y-u_t}_1
    \quad\text{for every }t\neq y,
    \label{eq:robust-condition}
\end{equation}
with certified radius
\begin{equation}
    \rho(x)=\min_{t\neq y}\frac{m_{y,t}(x)}{\norm{u_y-u_t}_1}.
    \label{eq:cert-radius}
\end{equation}
A competitor with $u_y=u_t$ and positive margin contributes an infinite radius by convention. There are no normalization branches and hence no branch-instability unknown outcome.

\subsection{Folded exact verification}

The final GPT-2-scale circuit is verified over an explicit finite domain, one prompt at a time. For each prompt, program-head selection is fixed by token identities and positions. LeakyReLU preactivation signs are certified externally in exact rational arithmetic; when a preactivation is exactly zero, both branches agree and the certificate records either branch as valid. The complete circuit then folds to exact rational candidate logits, or to an affine function of the final perturbation $\eta$ for robustness.

Functional equivalence and invariance reduce to exact ground comparisons. Edge necessity is checked by folding each zero-ablated circuit and recording witnesses. Robustness reduces to Equation~\eqref{eq:robust-condition}. The resulting formulas lie in linear real arithmetic and require no nonlinear search. As a soundness regression, the folded and legacy monolithic exact-rational encodings agree identically on eight anchor prompts.

\section{Verification-Oriented Transformer Components}

\subsection{Base architecture and primitive set}

We begin with the scaled dot-product attention architecture of Vaswani et al. and a GPT-style decoder-only Transformer \cite{vaswani2017attention,radford2019gpt2}. The preferred symbolic primitives are affine maps, comparisons, max/min, piecewise-linear activations, thresholding, fixed top-$k$ selection, and finite disjunctions over linear regions. These operations are readily expressible in linear real arithmetic with branch structure. Standard attention adds variable products, so the broader architecture is \emph{SMT-representable} rather than purely piecewise-affine.

\subsection{Signed L1 BandNorm and normalization removal}

LayerNorm divides by a data-dependent standard deviation \cite{ba2016layernorm}, introducing square roots and division. Signed L1 BandNorm instead centers an input $x\in\mathbb{R}^d$, splits positive and negative mass, controls each side through an L1 projection or bounded lift, recombines, recenters, and applies a learned affine map:
\begin{align}
    c &= x-\operatorname{mean}(x),\\
    p &= \max(c,0),\qquad n=\max(-c,0),\\
    z &= p'-n',\qquad z\leftarrow z-\operatorname{mean}(z),\\
    \BandNorm(x)&=\gamma\odot z+\beta.
\end{align}
The operator was the best-performing trainable normalization replacement tested from scratch. It is no longer the recommended large-scale route. Recent work shows that LayerNorm can be removed from GPT-2-family models after fine-tuning with little loss \cite{baroni2025layernormremoval}; Section~\ref{sec:ln-removal} extends removal to our sparsemax/LeakyReLU model and shows that it is both cheaper in loss and strictly easier to verify than BandNorm.

\subsection{Sparsemax attention and program heads}

Sparsemax replaces softmax with Euclidean projection onto the simplex \cite{martins2016sparsemax}. For $s\in\mathbb{R}^m$,
\begin{equation}
\sparsemax(s)=\argmin_{p\in\Delta^{m-1}}\norm{p-s}_2^2,
\qquad
\Delta^{m-1}=\left\{p\in\mathbb{R}^m:p_i\geq0,\ \sum_i p_i=1\right\}.
\end{equation}
Sparsemax is piecewise-linear in $s$ and produces exact zeros. Entmax provides a related family of sparse attention maps \cite{correia2019entmax}. However, standard sparsemax attention still computes $q_i^\top k_j$ and $\sum_j a_{ij}v_j$, both of which contain products of symbolic variables.

A \emph{program head} removes the first source of bilinearity. Its attention pattern is a frozen restricted-DSL program of token identities and positions rather than learned $W_Q/W_K$ scoring. The program produces a fixed selection or weighting rule for a concrete token sequence, while learned linear $W_V/W_O$ maps read and write residual features. After calibration, those maps are frozen. The retained circuit therefore contains no neural query-key product; for a concrete prompt the program selection is constant, and its value path is linear.

\subsection{LeakyReLU MLP}

GPT-2's GELU activation \cite{hendrycks2016gelu} is replaced with
\begin{equation}
\operatorname{LeakyReLU}(x)=
\begin{cases}
 x,&x\geq0,\\
 \alpha x,&x<0.
\end{cases}
\end{equation}
This activation is exactly piecewise-linear. The GPT-2 ablations in Section~\ref{sec:gpt2-prep} find essentially no OpenWebText cost from adding LeakyReLU to the sparsemax model.

\section{Small-Scale Verification}

\subsection{Tasks and direct verification}

The small model is an approximately 8K-parameter decoder-only Transformer with a custom 32-token vocabulary. It is trained jointly on two finite tasks:

\begin{itemize}[leftmargin=*]
    \item \textbf{Quote closing}: choose a single or double closing quote to match the unmatched opener.
    \item \textbf{Bracket type}: choose a square or curly closing delimiter to match the opener.
\end{itemize}

Each task has 128 exhaustive discrete inputs. Brute-force enumeration is sufficient for the finite functional-equivalence subclaim; that result primarily checks extraction and encoding. SMT adds the existential edge-witness queries and continuous residual-robustness quantification. Table~\ref{tab:small-direct} reports the original direct-verification results.

\begin{table}[H]
\centering
\small
\resizebox{\linewidth}{!}{%
\begin{tabular}{lrrrrrrrr}
\toprule
Task & Inputs & Edges & PyTorch & Equivalence & Invariance & Edge necessity & Robustness \\
\midrule
quote\_close & 128 & 3 & passed & verified & verified & verified & verified \\
bracket\_type & 128 & 6 & passed & verified & verified & verified & verified \\
\bottomrule
\end{tabular}%
}
\caption{Original small-model direct-verification results. Robustness uses $\epsilon_0=0.01$.}
\label{tab:small-direct}
\end{table}

\subsection{Branch-certificate incompleteness and norm-free remediation}

The successful checkpoint in Table~\ref{tab:small-direct} does not imply that BandNorm branch certificates are reliable across seeds. A matched retrain with the same intended architecture and $\epsilon_0=0.01$ returned \texttt{UNKNOWN\_BRANCH\_UNSTABLE} for both tasks. An $\epsilon$ sweep found branch adjacency even at $\epsilon_0/10=0.001$ and at every larger tested radius, with exactly 64 of 128 inputs branch-unstable per task. No decision violation was found. Crossing a BandNorm branch boundary does not imply a decision flip because BandNorm is continuous; it means only that the traced-branch proof is incomplete there.

A norm-free small model removes the branch certificate entirely. Table~\ref{tab:branch-remediation} shows the matched remediation. The original verified checkpoint remains valid; the new result identifies seed-dependent certifiability and motivates removal as the more reliable verification architecture.

\begin{table}[H]
\centering
\small
\begin{tabular}{lccccc}
\toprule
Checkpoint & $\epsilon=.001$ & $.0025$ & $.005$ & $.01$ & Decision violations \\
\midrule
Matched BandNorm retrain & unknown & unknown & unknown & unknown & 0 \\
Norm-free retrain & verified & verified & verified & verified & 0 \\
\bottomrule
\end{tabular}
\caption{Continuous-robustness remediation. Unknown denotes failure of traced branch stability, not a counterexample to the task decision.}
\label{tab:branch-remediation}
\end{table}

\subsection{Verified distillation at small scale}

We next replace retained attention heads with restricted programs. Each head is first isolated by per-head extraction and block regression. Program synthesis is accepted on exact projected agreement over the declared finite domain, not on equality of the complete attention matrix. The accepted programs nevertheless recover related sparse supports: support Intersection-over-Union ranges from 0.69 to 0.88, with mean sparsemax support 1.33 positions versus 3.5 under a softmax counterfactual.

Installing a program does not by itself make it causal. Three progressively stronger objectives reveal this:

\begin{enumerate}[leftmargin=*]
    \item \textbf{Plain healing} restores task behavior but silently routes around the program, even at this 8K-parameter scale; one run reaches KL $7.4\times10^{-6}$ while the program remains bypassable.
    \item \textbf{Ablation-aware healing} makes the replaced heads individually important, but the mechanism migrates through a newly retained downstream head.
    \item \textbf{Core-aware healing} applies joint and individual suppression in both the full model and the preregistered circuit. One bounded chase round replaces the new head as well and re-extracts the circuit.
\end{enumerate}

The final quote and bracket circuits contain no active neural-attention bilinear terms and verify all four properties. Across the small task artifacts, the parameterized QK footprint falls from 1,088 to 272; within the retained verified paths, no active neural QK terms remain. The result also lowers encoding cost. We attribute costs per edge rather than comparing raw totals across different topologies.

\begin{figure}[H]
    \centering
    \includegraphics[width=0.72\linewidth]{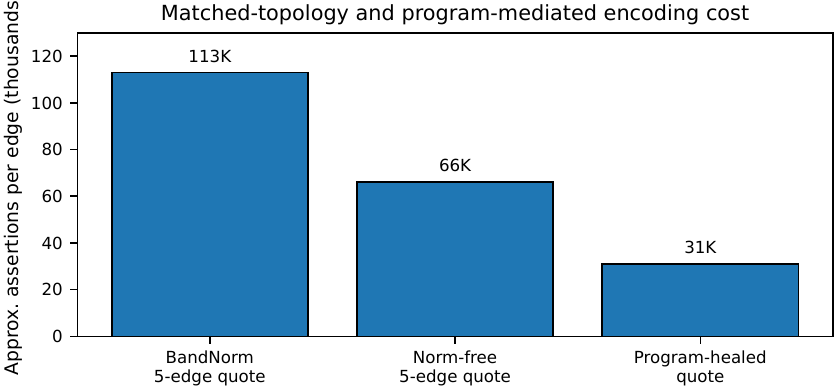}
    \caption{Verification encoding cost. The BandNorm and norm-free bars use a matched five-edge quote topology. The program-healed circuit is separately re-extracted, so only the normalized per-edge quantity is compared.}
    \label{fig:encoding-cost}
\end{figure}

\begin{table}[H]
\centering
\small
\resizebox{\linewidth}{!}{%
\begin{tabular}{lccc}
\toprule
Circuit variant & Approx. assertions/edge & Active neural QK terms & All four verified \\
\midrule
BandNorm, five-edge quote & 113K & yes & no (branch unknown) \\
Norm-free, five-edge quote & 66K & yes & yes \\
Program-healed quote & 31K & no & yes \\
\bottomrule
\end{tabular}%
}
\caption{Matched and normalized encoding-cost accounting. BandNorm contributes roughly 50 assertions per normalization instance; bilinear elimination approximately halves per-edge cost again after normalization removal.}
\label{tab:encoding-cost}
\end{table}

\section{GPT-2-Scale Model Preparation}
\label{sec:gpt2-prep}

The small experiments establish proof mechanics. We next prepare a GPT-2-small-scale model for circuit verification. All variants use the same tokenizer, context length, OpenWebText training pipeline \cite{gokaslan2019openwebtext}, and local evaluation code; WikiText-103 provides the secondary validation benchmark \cite{merity2016pointer}. The local GPT-2 baseline, rather than the separately evaluated pretrained checkpoint, is the primary comparison anchor.

\subsection{Trainability ablations}

The local baseline reaches OpenWebText validation loss 3.1340 and WikiText-103 perplexity 52.9820. Table~\ref{tab:gpt2-results} adds two results to the earlier ablations: a sparsemax+LeakyReLU model with standard LayerNorm, and the post-fold norm-free endpoint derived from it.

\begin{table}[H]
\centering
\small
\resizebox{\linewidth}{!}{%
\begin{tabular}{lrrrr}
\toprule
Model & OWT loss & $\Delta$ OWT & WT103 ppl & $\Delta$ WT103 \\
\midrule
Local GPT-2 baseline & 3.1340 & 0.0000 & 52.9820 & 0.0000 \\
Sparsemax only & 3.1973 & +0.0633 & 55.7227 & +2.7407 \\
Sparsemax + LeakyReLU + LayerNorm & 3.1969 & +0.0629 & 57.1855 & +4.2035 \\
Signed L1 BandNorm only & 3.3180 & +0.1840 & 61.8900 & +8.9080 \\
BandNorm + sparsemax + LeakyReLU & 3.3300 & +0.1960 & 62.1100 & +9.1280 \\
Norm-free post-fold & 3.2056 & +0.0716 & 52.5124 & $-0.4696$ \\
\bottomrule
\end{tabular}%
}
\caption{GPT-2-scale trainability results. The norm-free model is derived from the sparsemax+LeakyReLU+LayerNorm source by a separate 5,000-step removal run.}
\label{tab:gpt2-results}
\end{table}

\begin{figure}[H]
\centering
\begin{subfigure}{0.49\linewidth}
    \centering
    \includegraphics[width=\linewidth]{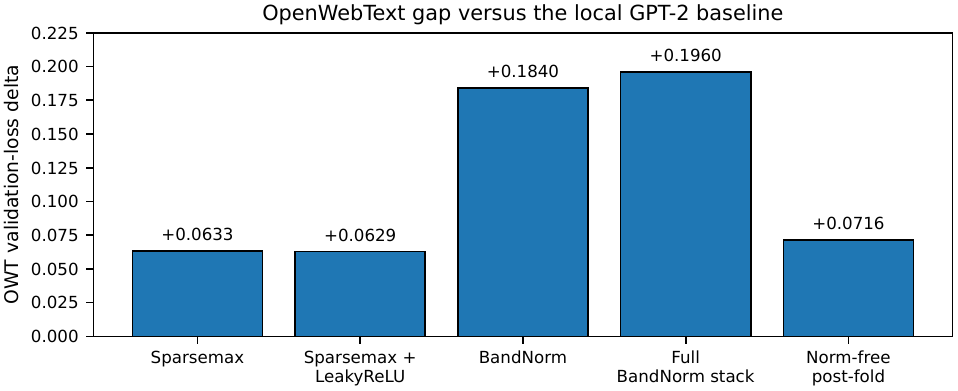}
\end{subfigure}
\begin{subfigure}{0.49\linewidth}
    \centering
    \includegraphics[width=\linewidth]{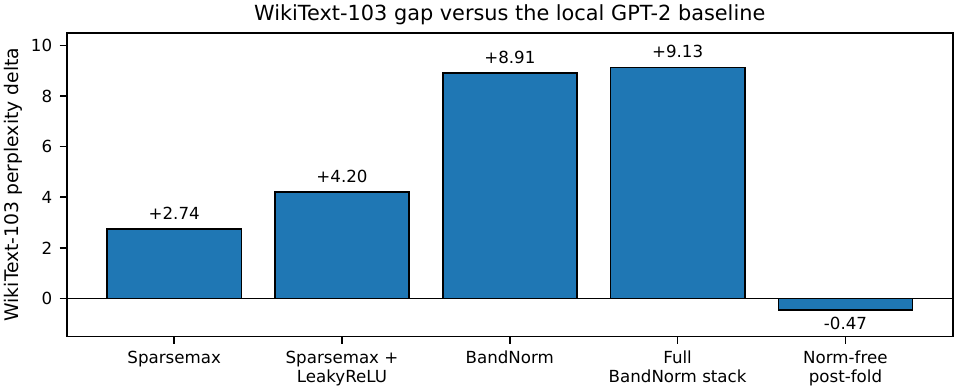}
\end{subfigure}
\caption{GPT-2-scale loss gaps relative to the local baseline. Sparsemax and LeakyReLU account for little OpenWebText degradation; BandNorm is the main from-scratch performance bottleneck. The WikiText improvement of the norm-free run is a single-run observation and may partly reflect its additional 5,000 steps.}
\label{fig:gpt2-deltas}
\end{figure}

The sparsemax-only and sparsemax+LeakyReLU runs are nearly identical on OpenWebText, isolating essentially no measured LeakyReLU cost there. BandNorm produces the largest gap. This makes normalization removal, rather than a new normalization operator, the preferred path for the scale experiment.

\subsection{LayerNorm removal at GPT-2 scale}
\label{sec:ln-removal}

Starting from the sparsemax+LeakyReLU+LayerNorm checkpoint, we fine-tune for 5,000 optimizer steps while sequentially replacing all 25 normalization instances with fixed-standard-deviation affine maps. The schedule reaches the fully attenuated endpoint at step 3,400, leaving 1,600 endpoint steps and processing approximately 1.311 billion tokens in total. Every affine map is then folded into its consumer.

The post-fold OpenWebText loss is 3.2056, only $+0.0087$ relative to its LayerNorm source and 0.1124 better than the BandNorm-only gate. This extends LayerNorm-removal results \cite{baroni2025layernormremoval} to a sparsemax model. On WikiText-103, the folded model reaches perplexity 52.5124, improving by 4.6731 over its LayerNorm source and by 0.4696 over the local baseline. The latter is a single-run result and may partly reflect the additional fine-tuning; it is not presented as a generalization claim.

An initially large fold discrepancy was a BF16 comparison artifact: algebraically equivalent graphs rounded differently under autocast. Repeating the fold in FP64 and comparing in FP32 gives maximum absolute logit error $6.58\times10^{-5}$, relative L2 error $9.09\times10^{-7}$, top-1 agreement 1.000, and validation-loss delta $3.67\times10^{-5}$. Verification-path forwards use FP32 and exports use exact rational arithmetic.

The model of record for subsequent work is therefore norm-free. BandNorm remains an informative from-scratch negative result: it trains, but it loses more language-modeling performance and introduces branch-certification incompleteness that normalization removal eliminates by construction.

\subsection{Provability gained by removal}

Removing the final norm converts the task decision interface to an affine unembedding plus argmax. It eliminates BandNorm branch formulas, makes the unknown robustness status structurally unreachable, and yields the exact radius in Equation~\eqref{eq:cert-radius}. Removal also reduces encoding size at small scale. It does not solve the remaining GPT-2 bottleneck: neural QK attention still contributes bilinear terms at width 768. That bottleneck motivates the localization study and constructive program-head route below.

\section{Discovery-Based Localization at GPT-2 Scale}

\subsection{Declared protocols and untouched gates}

The pilot GPT-2 behavior dataset had a defect: its nominal 128 rows consisted of 16 unique templates repeated eight times. Those records are retained for provenance but are not used as evidence for generalization. We then ran three deterministic, model-independent gated protocols. Each protocol burns every previously viewed gate as development data and evaluates the selected circuit once on a fresh untouched gate.

\begin{table}[H]
\centering
\small
\resizebox{\linewidth}{!}{%
\begin{tabular}{lrrrrl}
\toprule
Protocol & Development used for selection & Fresh gate & Quote circuit & Bracket circuit & Outcome \\
\midrule
Pilot & 16 unique templates repeated $8\times$ & none & nominally exact & nominally exact & defective domain \\
A & 256 synthesis prompts & 256 & exact & 247/256 & stop before healing \\
B & 512 burned prompts & 256 & exact & 255/256 & stop before healing \\
C & 768 burned prompts & 512 & 511/512 (17 edges) & 512/512 (340 edges) & final stop \\
\bottomrule
\end{tabular}%
}
\caption{Localization protocols. Protocol C's bracket circuit is exact only by retaining 158 nodes and all 144 attention heads. Exact gate outcomes are behavioral measurements under zero ablation, not formal proofs.}
\label{tab:localization}
\end{table}

\begin{figure}[H]
    \centering
    \includegraphics[width=0.88\linewidth]{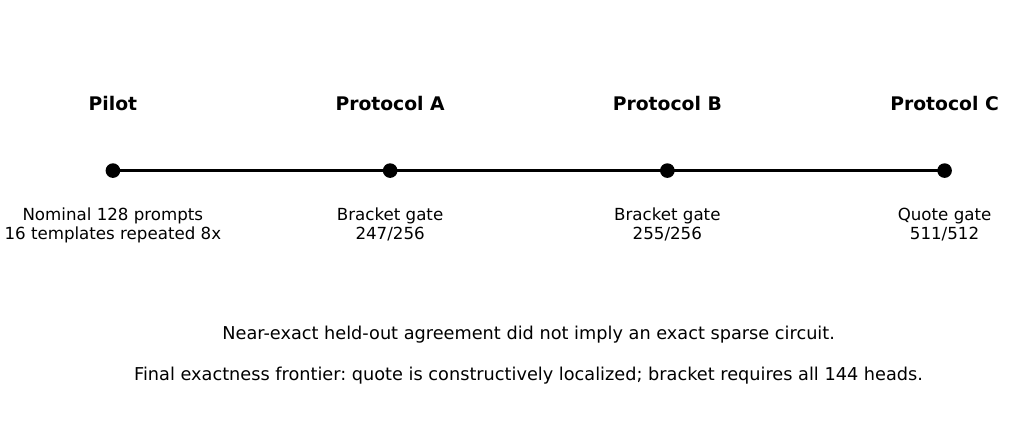}
    \caption{Protocol progression and the measured sparsity--exactness frontier. Every failed gate is burned and never reused as untouched evidence.}
    \label{fig:protocol-timeline}
\end{figure}

Protocol A selects on 256 prompts and finds a bracket circuit exact on development but only 247/256 on its fresh gate. Protocol B burns both Protocol A splits, selects on 512 prompts, and reaches 255/256 on a new gate. Protocol C burns all 768 viewed prompts, uses a preregistered worst-case-margin-first selection rule, and evaluates on 512 new prompts. The 17-edge quote circuit misses one double-quote prompt. The bracket circuit reaches 512/512 only after re-extraction expands to 340 edges, 158 nodes, and all 144 heads.

\subsection{A localization frontier, not a task-difficulty result}

Quote closing and bracket type implement the same high-level operation: detect an opener and copy its type to the projected closing-token decision. They differ primarily in alphabet. The sharp difference in sparse localization therefore cannot be explained simply by algorithmic complexity. It is a property of how the dense model distributes the two behaviors. The bracket task is related to bounded bracket-language analyses in prior work \cite{hewitt2020dyck,yao2021dyck}, though our projected delimiter task is much narrower than full Dyck-language recognition or generation.

The held-out sequence establishes a negative result for discovery-based exactness: sparse zero-ablation circuits can be repeatedly near-exact while failing on one or a few untouched inputs. We do not select again against Protocol C's gate. Instead, the construction experiment declares a different bounded claim.

\subsection{The declared construction domain}

The construction domain $D$ is the ordered union of the frozen Protocol C development and gate manifests: 1,280 unique prompts, balanced 640/640 between quote classes and pinned by prompt-set hashes in the public evidence release. There is no held-out split and no claim that the resulting artifact generalizes beyond $D$. This is deliberate. The goal is to construct and verify an exact circuit on a declared domain after measuring why post-hoc sparse extraction did not produce one.

\section{Verified Distillation at GPT-2 Scale}

\subsection{Program synthesis and the limit of healing}

The selected quote circuit initially retains attention heads 7.11 and 9.0. For each head, a search enumerates restricted programs over token identities and positions. Acceptance requires exact projected agreement on $D$ for the relevant circuit behavior. Accepted programs replace the neural $W_Q/W_K$ mechanism at inference; their linear readouts remain available for calibration.

Before program replacement, ablating the selected circuit produced a useful inadvertent lesion study. Language-model quality remained within the locked budget (OpenWebText perplexity $25.46\leq 28.6176$), while projected agreement fell to $0.9375$ in the full forward and $0.75$ in the selected-circuit forward. This selective task damage at intact aggregate perplexity is causal evidence that the extraction contains task-relevant computation, although it does not establish completeness or uniqueness.

Whole-model healing is not sufficient to establish that the programs implement the behavior. The preregistered core-aware route failed exact agreement after two 10,000-step attempts. An engineering continuation corrected an accumulation bug and added full-forward supervision, reaching exact behavior within the perplexity budget, but the programs were jointly removable without changing the full-model task decision. Suppression losses reduced margins without forcing an argmax change. This is a scale limit of mechanism-pinning by objective alone.

\subsection{Causal audit}

A pre-healing audit resolves whether the bypass was created by training. In the untouched norm-free model, jointly zero-ablating heads 7.11 and 9.0 leaves 1,256/1,280 projected decisions correct. Ablating the entire selected circuit leaves 710/1,280. Installing the programs without training produces the identical 1,256/1,280 joint-lesion result. All 23 zero-step full-model errors lie within the 24 prompts not covered by the native backup route.

\begin{figure}[H]
    \centering
    \includegraphics[width=0.85\linewidth]{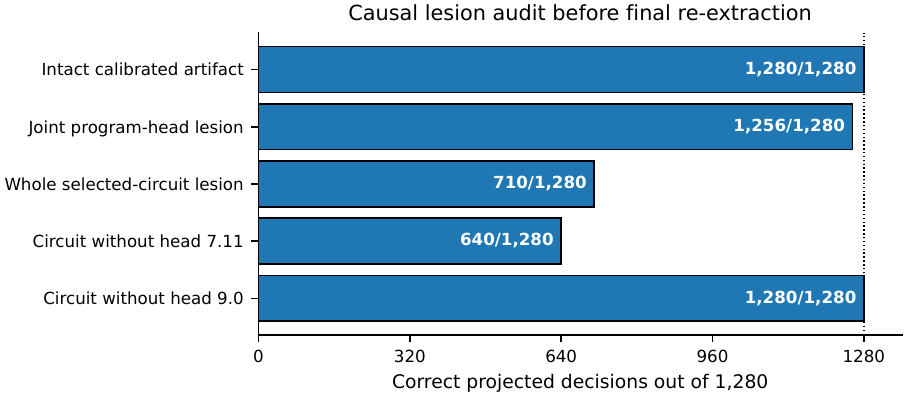}
    \caption{Causal lesion audit. Counts are measured projected decisions, not formal properties. The joint program-head and whole-circuit lesions characterize native redundancy; individual head lesions motivate the final re-extraction.}
    \label{fig:lesion-audit}
\end{figure}

Thus the redundant path pre-exists at 98.1 percent coverage. Healing did not invent it; healing closed the remaining 24 inputs. Figure~\ref{fig:lesion-audit} summarizes the lesion measurements, including the later finding that head 9.0 is individually redundant while head 7.11 is strongly necessary.

\subsection{Causally constrained calibration}

We abandon whole-model healing and freeze every non-program parameter, recording hashes before and after calibration. Only parameters local to the program heads are trainable. The escalation ladder permits: (i) two scalar output gains, (ii) per-channel diagonal gains, and (iii) program-local $W_V/W_O$. Every trainable contribution disappears under program lesion, so the lesioned forward is bit-identical to the zero-step baseline in parameters, decisions, and margins. This resembles parameter-efficient fine-tuning mechanically \cite{houlsby2019adapters}, but the frozen parameter set plays a proof role rather than merely reducing training cost.

\begin{table}[H]
\centering
\small
\begin{tabular}{lrrrr}
\toprule
Calibration rung & OWT ppl & Full agreement & Circuit agreement & Joint-head lesion \\
\midrule
Scalar gains & 25.6691 & 1280/1280 & 1280/1280 & 1256/1280 \\
Diagonal gains & 25.7157 & 1280/1280 & 1280/1280 & 1256/1280 \\
Program-local $W_V/W_O$ & 25.5707 & 1280/1280 & 1280/1280 & 1256/1280 \\
\bottomrule
\end{tabular}
\caption{Constrained-calibration ladder. Every rung remains within the locked perplexity budget 28.6176 and leaves the whole-circuit lesion at 710/1280. The $W_V/W_O$ rung is the model of record.}
\label{tab:calibration}
\end{table}

All three rungs pass exact full and circuit-only agreement and the locked perplexity budget. The $W_V/W_O$ rung has the best perplexity and is the model of record. Head 9.0 remains individually redundant: removing it changes no projected decision, whereas removing 7.11 drops circuit-only accuracy to 640/1,280. Because the no-new-bypass guarantee is now deductive, the preregistered individual-head gate is amended transparently: final causality is evaluated by the lesion identity, joint program-set necessity, whole-circuit necessity, and formal circuit-internal edge necessity. Calibration should not manufacture non-redundancy that the frozen model did not possess.

\subsection{The verified circuit}

Re-extraction on all of $D$ removes head 9.0 and selects the three-edge circuit
\begin{equation}
    \text{embedding}\ \longrightarrow\ \text{MLP 0}\ \longrightarrow\ \text{program head 7.11}\ \longrightarrow\ \text{projected logits}.
\end{equation}
Its coarse structure independently matches the string-closing motif in the weight-sparse circuits of Gao et al. \cite{gao2025weightsparse}: an early MLP constructs quote features and one later head copies the type. This is a structural comparison, not an identity claim.

\begin{figure}[H]
    \centering
    \includegraphics[width=0.98\linewidth]{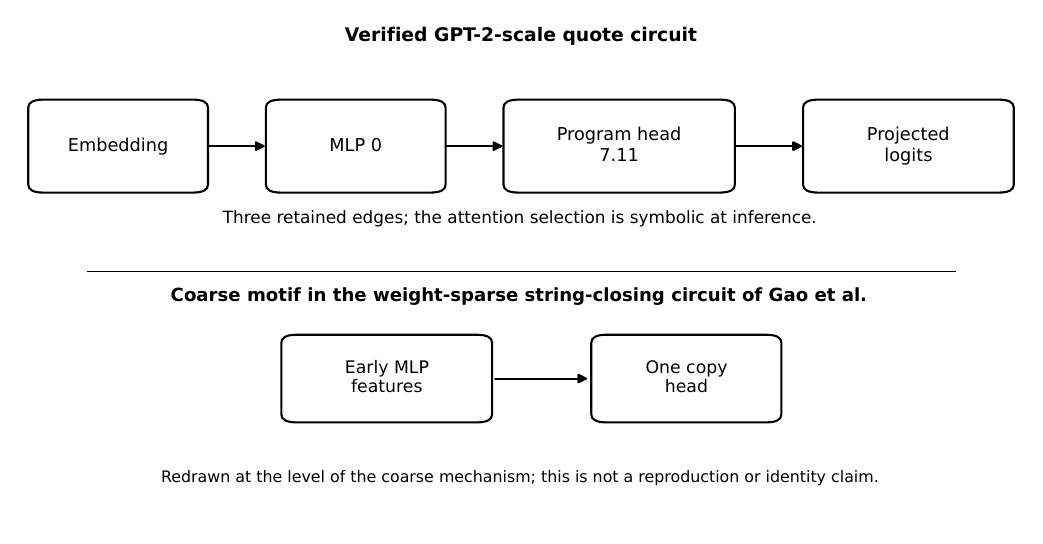}
    \caption{The final verified quote circuit and a redrawn coarse structural comparison to the string-closing motif in weight-sparse models.}
    \label{fig:verified-circuit}
\end{figure}

The exact encoder matches PyTorch candidate logits to maximum error $1.11\times10^{-8}$. The legacy monolithic encoding builds one input in approximately 494 seconds and validates 9,216 LeakyReLU trace constraints as satisfiable. The folded verifier certifies signs in exact rational arithmetic, contracts the circuit once, and folds all 1,280 inputs.

\begin{table}[H]
\centering
\small
\resizebox{\linewidth}{!}{%
\begin{tabular}{ll}
\toprule
Quantity & Verified or measured result \\
\midrule
Declared domain & 1,280 unique hash-pinned prompts, balanced 640/640 \\
Circuit & 3 edges: embedding $\rightarrow$ MLP 0 $\rightarrow$ program head 7.11 $\rightarrow$ logits \\
Encoder agreement & max candidate-logit error $1.11\times10^{-8}$ \\
Projected functional equivalence & verified, 1280/1280 \\
Content invariance & verified, 1280/1280 \\
Edge necessity & verified, 640 exact witnesses for each of 3 edges \\
Continuous robustness & verified at $\epsilon=0.01$ \\
Certified radius distribution & min 0.01514986; median 0.02260854; max 0.03640277 \\
Verification time & 66.61 s total; independently rounded components: 16.04 s contraction + 50.58 s folds + 0 measured solver-search s \\
Model quality & OWT ppl 25.5707 vs. 24.6703 norm-free base; budget 28.6176 \\
Parameter integrity & every non-program parameter hash-identical to the base \\
\bottomrule
\end{tabular}%
}
\caption{Final GPT-2-scale verified-distillation result. Formal claims apply to the calibrated artifact over the declared domain. Perplexity, runtime, and floating-point encoder agreement are measured supporting evidence.}
\label{tab:gpt2-verified}
\end{table}

\begin{figure}[H]
    \centering
    \includegraphics[width=0.72\linewidth]{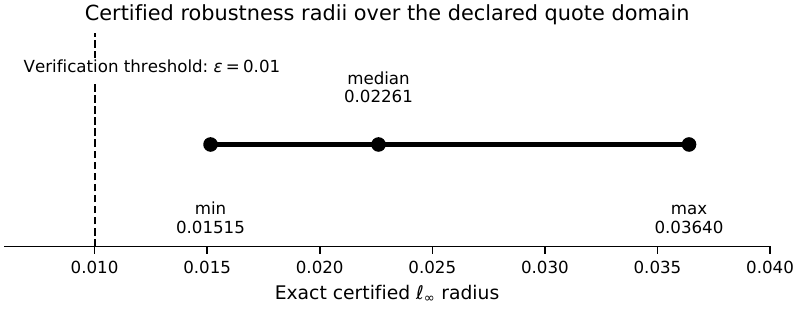}
    \caption{Exact certified robustness-radius summary over the 1,280-prompt quote domain. The complete per-input records are included in the evidence release.}
    \label{fig:radii}
\end{figure}

Every edge is load-bearing for one complete input class, yielding 640 witnesses per edge. Robustness at the locked $\epsilon=0.01$ follows from exact margins and Equation~\eqref{eq:cert-radius}; Figure~\ref{fig:radii} shows the range. Because the circuit is norm-free and its attention is programmatic, there are neither normalization branches nor neural bilinear QK terms, and the unknown verification outcome is structurally unreachable.

One solver counterexample on a legacy maximum-length-3 synthetic input is preserved outside $D$. It is a genuine boundary witness, not a failed in-domain claim.

\subsection{What is and is not verified}

\noindent\fbox{\begin{minipage}{0.94\linewidth}
\small
\textbf{Verified.} The four properties of Table~\ref{tab:properties} for the three-edge circuit in the calibrated model of record over the declared quote domain $D$; exact-rational agreement of folded and monolithic encodings on anchors; and the lesion identity induced by the frozen/trainable parameter partition.

\medskip
\textbf{Measured but not formally verified.} OpenWebText and WikiText perplexity, lesion accuracies of the full model, encoder-to-PyTorch floating-point error, support IoU, held-out localization gates, and runtime.

\medskip
\textbf{Not claimed.} That the original model's neural attention implements the program; that the unmodified model has a three-edge native circuit; correctness outside $D$; GPT-2-scale verification for bracket type; full-vocabulary equivalence; or end-to-end verification of the language model.
\end{minipage}}

\section{Surrogate-Mediated Symbolic Circuit Verification}

Verified distillation makes the symbolic program part of the artifact. A complementary route leaves the teacher circuit unchanged and uses an external surrogate. Given a bounded teacher relation $f_{\mathrm{circuit}}:X\rightarrow Y$, we fit a small ReLU network $S$, synthesize a program $P$ from a restricted DSL, and ask
\begin{equation}
    \exists x\in D:\ P(x)\neq d_T(S,x).
\end{equation}
The researcher supplies the task domain and DSL; the method does not discover the task from scratch. Its purpose is to turn a proposed algorithmic explanation into a refutable statement.

The experiments use quote tracking and bracket counting artifacts derived from the sparse-circuit work of Gao et al. \cite{gao2025weightsparse}. ``Finite-dataset agreement'' means agreement on examples used to fit or select the program, while surrogate-teacher agreement is evaluated over the bounded abstract domain used for verification.

\begin{table}[H]
\centering
\small
\resizebox{\linewidth}{!}{%
\begin{tabular}{lrrll}
\toprule
Task & Finite-dataset agreement & Surrogate-teacher agreement & Program equivalence & Outcome \\
\midrule
Quote & 100\% & 100\% & refuted for restricted template & counterexample returned \\
Bracket & 100\% & 100\% & verified & compact program certified \\
\bottomrule
\end{tabular}%
}
\caption{Surrogate-mediated symbolic verification. Dataset-level agreement is not sufficient for bounded formal equivalence.}
\label{tab:scd-results}
\end{table}

For the quote task, the solver returns the abstract input $[2,0,2]$, rendered as \texttt{[!, ", !]}. The teacher and surrogate choose a single quote while the proposed program chooses a double quote. For the bracket task, the synthesized program is equivalent throughout the bounded abstract domain. The quote counterexample and the GPT-2 outside-domain counterexample instantiate the same proof-or-counterexample discipline at different stages: one refutes an explanation inside its proposed domain, while the other records the boundary of a deliberately narrower verified claim.

This surrogate route is weaker than verified distillation because the proof concerns $S$ or a finite teacher relation. It remains useful when modifying the model is undesirable or when a symbolic hypothesis should be tested before architectural intervention.

\section{Discussion}

\subsection{At scale, construction replaces post-hoc discovery}

The scale result is not that a dense GPT-2 circuit became easy to encode. Naive encodings of extracted neural circuits remain intractable because width-768 QK and attention-value products create nonlinear formulas. The successful route changes the artifact in two targeted ways: remove normalization, and replace the retained attention mechanism with a restricted program. Constrained calibration then restores exact behavior without permitting a newly trained bypass. In this sense, we must build the object we can verify.

\subsection{Sparse explanation and exactness are different goals}

The three gated protocols show that a circuit can be exact on all extraction data and fail on one untouched input. Conversely, exactness may be recovered by retaining an effectively unpruned graph. Reporting only average agreement or a selected development domain obscures this frontier. The result argues for explicit gates, burned evaluation data, worst-case margins, and a clear distinction among sparse faithfulness, bounded exactness, and verified construction.

\subsection{Redundancy is not the same as bypass}

The causal audit finds a native route that already covers 1,256/1,280 quote prompts when the selected heads are removed. Whole-model healing cannot distinguish creating a bypass from exploiting one that is already present. The frozen calibration design separates the two: training cannot change the lesioned forward, while the native route remains visible. This does not eliminate redundancy; it makes its provenance auditable.

\subsection{Why bounded verification remains useful}

Enumeration is enough to compare discrete decisions on 1,280 listed prompts. The verification framework adds more than that comparison: existential edge witnesses, exact invariance relations, continuous perturbation radii, explicit counterexamples, and an auditable chain from exported weights to exact arithmetic. The scope remains bounded. The right interpretation is a precise certificate for a declared artifact and domain, not evidence of arbitrary-length generalization.

\section{Related Work}

\paragraph{Mechanistic circuits and sparse models.}
Induction-head analysis and grokking studies show that algorithmic behaviors can be localized and interpreted \cite{olsson2022induction,nanda2023progress}. ACDC automates edge pruning under ablation-based objectives \cite{conmy2023acdc}. Gao et al. train weight-sparse Transformers and release unusually legible circuits for synthetic tasks \cite{gao2025weightsparse}. Our GPT-2 result is complementary: the base model is densely trained, and a task circuit is made verifiable through program replacement and causally constrained calibration.

\paragraph{Programmatic attention.}
Hayes et al. synthesize Python programs that approximate attention patterns in GPT-2-family models and show that programmatic surrogates can replace a substantial fraction of heads with moderate perplexity cost \cite{hayes2026programattention}. Their acceptance criteria are aggregate and behavioral, including attention-support overlap and model quality. Our programs use a deliberately restricted SMT-native DSL, require exact projected agreement on a declared domain, and are integrated with a deductive no-new-bypass calibration rule and formal circuit verification. The two approaches share the premise that executable programs can replace opaque attention, but optimize for different guarantees.

\paragraph{Normalization and sparse attention.}
LayerNorm stabilizes hidden-state dynamics \cite{ba2016layernorm}. Baroni et al. show that it can be removed from GPT-2-family models after fine-tuning \cite{baroni2025layernormremoval}; we extend that route to sparsemax attention and quantify the verification benefit. Sparsemax and entmax replace dense softmax distributions with sparse alternatives \cite{martins2016sparsemax,correia2019entmax}.

\paragraph{Neural-network verification.}
Z3 supports satisfiability modulo combinations of logical theories \cite{demoura2008z3}. Reluplex and Marabou develop SMT-style verification for ReLU networks \cite{katz2017reluplex,katz2019marabou}. Our target differs from standard adversarial robustness of feed-forward classifiers: the objects are extracted Transformer circuits, the output is a task projection, and the claims include edge necessity and symbolic program equivalence.

\paragraph{Parameter-efficient adaptation.}
Adapter methods freeze a pretrained backbone and train a small parameter subset \cite{houlsby2019adapters}. Constrained calibration is mechanically parameter-efficient, but its distinguishing role is logical: the trainable set is chosen so that every trainable contribution vanishes under a specified lesion, giving Equation~\eqref{eq:lesion-identity}.

\paragraph{Bounded hierarchical languages and invariant aggregation.}
Dyck-language analyses characterize bounded hierarchical processing in recurrent and self-attention networks \cite{hewitt2020dyck,yao2021dyck}. Deep Sets characterize permutation-invariant functions under suitable assumptions \cite{zaheer2017deepsets}; Appendix~\ref{app:deepsets} uses this view to motivate attention operators without bilinear QK.

\section{Limitations}

\begin{enumerate}[leftmargin=*]
    \item All formal claims are bounded to declared domains and candidate-token projections.
    \item The GPT-2-scale verified object is a constructed calibrated artifact, not the original model's neural mechanism.
    \item Only quote closing is verified at GPT-2 scale. Bracket type is reported as a localization boundary; the planned bracket program graft is not included.
    \item Native redundancy coexists with the verified circuit. It is frozen and measured, not removed.
    \item The declared-domain experiment is not a held-out-generalization result. The preserved outside-domain counterexample demonstrates this boundary.
    \item The norm-free WikiText improvement and the BandNorm branch-adjacency result are single-run or per-seed observations.
    \item Verification concerns the exported exact-rational representation. FP32 encoder checks connect it to execution, but do not prove bit-for-bit equivalence to every hardware floating-point path.
    \item Folded verification depends on fixed program behavior and exact sign certificates. A different prompt domain requires new folds and certificates.
    \item The program DSL is restricted. Failure to find a program does not imply that no symbolic explanation exists.
    \item Edge necessity is relative to zero ablation and the selected graph; it is not global circuit minimality.
    \item Language-model quality is measured by perplexity only. The calibrated artifact is not evaluated on a broad downstream benchmark suite.
\end{enumerate}

\section{Future Work}

The immediate next experiment is a constructed bracket circuit using the same program-head and constrained-calibration recipe. It is intentionally excluded from the present result regardless of its eventual outcome. More broadly, weight-sparse pretraining may improve the localization stage before verification; the released sparse circuits of Gao et al. offer a natural gate survey, and a fused sparse-pretraining/program-distillation experiment would test whether exact circuits can be obtained with less intervention.

The program DSL should also expand beyond simple positional and token-identity rules. Scan, match, and bounded-stack primitives are natural hypotheses for richer syntax tasks. Counterexamples can drive this expansion automatically: a refuted program suggests a new predicate or state variable rather than an informal explanation patch.

A second architectural direction is attention without bilinear QK at the full-model level, not merely inside a selected circuit. Hard selection, Boolean gates, and Deep Sets-style aggregators with tractable coefficients could preserve the algorithmic role of attention while admitting linear or mixed-integer verification. Because verified distillation never encodes the frozen neural remainder, it may also apply to ordinary softmax-pretrained models; testing that decoupling is an important practical direction.

Finally, diffuse mechanisms may require relocation rather than simple readout calibration: fine-tuning that moves a behavior into a declared symbolic module while preserving an explicit lesion identity. Any such method should retain the central claim discipline of this paper: specify the verified object, freeze the domain, and distinguish measured behavior from formal proof.

\section{Conclusion}

Verifiable Transformers turns mechanistic explanations into bounded claims with explicit proof objects. Small circuits can be encoded directly; hard-to-encode circuits can be tested through external surrogates; and at GPT-2 scale, a selected mechanism can be replaced by a symbolic program and calibrated under a deductive no-new-bypass constraint. The resulting three-edge quote circuit verifies functional equivalence, invariance, edge necessity, and continuous robustness over a hash-pinned 1,280-prompt domain. The result does not verify unmodified GPT-2. It demonstrates a constructive path from circuit discovery to a model artifact whose mechanism is simple enough to prove.

\appendix

\section{Notation and Property Details}
\label{app:notation}

\begin{longtable}{p{0.19\linewidth}p{0.72\linewidth}}
\toprule
Symbol & Meaning \\
\midrule
$M$ & trained model before task-specific intervention \\
$\widetilde M$ & program-installed, constrained-calibration artifact \\
$P$ & symbolic reference program returning a token in $T$ \\
$D$ & declared bounded domain \\
$T$ & task-specific candidate-token set \\
$F_T(x)$ & candidate logits of model or circuit $F$ \\
$d_T(F,x)$ & projected decision $\argmax_{t\in T}F_t(x)$ \\
$C_E$ & extracted circuit with retained edge set $E$ \\
$C_{E\setminus\{e\}}$ & circuit after deleting retained edge $e$ \\
$r_E(x)$ & final residual before final normalization, if present \\
$G_T(r)$ & projected decision after the final interface and unembedding \\
$H$ & restricted symbolic attention program \\
$L_H$ & lesion that zeroes program-head contributions \\
$\rho(x)$ & exact certified final-residual robustness radius \\
\bottomrule
\end{longtable}

For BandNorm models, robustness uses a traced branch certificate. The verifier first attempts to prove that the branch remains fixed throughout the perturbation set and then proves that no decision flip is possible inside that branch. Failure of the first step returns unknown. For norm-free models, Equation~\eqref{eq:cert-radius} replaces the branch procedure.

\section{Signed L1 BandNorm and Remediation Details}
\label{app:bandnorm}

Signed L1 BandNorm was motivated by a sequence of failed normalization replacements. Affine-clamp-affine transformations lacked centering and scale control. MAD-based piecewise-linear normalization introduced discontinuous gain buckets. Unbounded leaky clamps preserved gradients but allowed residual growth. Bounded elementwise clamps stabilized training but destroyed dynamic range. Projection-based signed mass control performed best among the from-scratch replacements.

The matched remediation separates two questions: whether a checkpoint is robust and whether the traced proof can certify it. The branch-adjacent BandNorm retrain has no found decision counterexample throughout the tested radii, but the verifier cannot reason across the branch transition. The norm-free model proves the decision directly. Exactly half of each task domain being branch-adjacent suggests input-class structure rather than random numerical noise, but this observation is seed-specific.

Cost accounting attributes approximately 50 assertions to each BandNorm instance. We do not compare raw assertion totals across different circuit topologies. The 113K-versus-66K comparison uses the same five-edge quote topology; the 31K program-healed result is normalized per edge after re-extraction.

\section{Small-Scale Program Distillation Details}
\label{app:small-distill}

The restricted DSL describes attention supports using token identities, relative or absolute positions, fixed offsets, and Boolean combinations. Candidate programs are evaluated first at the isolated head and then through block regression. Exact projected decision agreement is the acceptance gate. Attention-map overlap is diagnostic only, because multiple attention maps can induce the same task computation.

The healing progression is retained as a negative result. Plain task loss can make a program observationally irrelevant. Individual ablation pressure can move the mechanism to a new head. Core-aware joint and individual lesions constrain both full and circuit-only forwards, but may still require a bounded chase when re-extraction exposes a new path. The final small circuits pass migration, contain no active neural QK products, and verify all four properties.

\section{LayerNorm Removal and Numerical Precision}
\label{app:ln-removal}

The removal run processes a nominal 1.311 billion tokens: 5,000 steps, effective global batch 256, sequence length 1,024. Fixed-standard-deviation attenuation completes at step 3,400. Measured frozen standard deviations increase with depth, from approximately 0.031 in early layers to 0.303 near the output, answering the concern that a single fixed scale would be appropriate everywhere.

Folding changes operation order. Under BF16 autocast this can produce apparently large elementwise differences even when the two graphs are algebraically equivalent. The accepted validation therefore performs the fold in FP64, executes comparison forwards in FP32, and applies fail-closed thresholds to maximum absolute error, relative L2 error, top-1 agreement, and evaluation loss. Exact-rational exports are produced only after that check.

\section{Protocol Generations and Domain Construction}
\label{app:protocols}

Each behavior-domain generator is deterministic and independent of model outputs. Manifests record prompt hashes, tokenizer-vocabulary hashes, candidate-token identities, contextual opener alignment, template coverage, opener positions, and token-length ranges. A gate is evaluated only after circuit and program selection, and a viewed gate is permanently burned.

The final gated protocol changes selection from sparsity-first to worst-case-margin-first, then edge count, then threshold. This rule was preregistered in response to repeated one-example bracket failures near the circuit boundary. The protocol still stops because the quote circuit misses one prompt. The exact bracket result is reported with its full size rather than treated as a sparse success.

The declared construction domain $D$ is not another gate. It is the frozen union of all 1,280 final-protocol development and gate prompts. Programs, calibration, re-extraction, and verification all use this domain, and the paper makes no held-out claim for the constructed artifact.

\section{Folded Verification Soundness}
\label{app:folded}

For each prompt, the verifier evaluates embeddings and the frozen program selection exactly. Each LeakyReLU preactivation is converted to a rational number. A positive or negative sign selects the corresponding affine branch. At zero, both branches return zero, so either branch is sound. Substitution yields exact rational candidate logits.

For invariance, the verifier compares declared prompt pairs after folding. For edge necessity, it repeats the contraction under each edge lesion and records all disagreement witnesses. For robustness, it retains the final perturbation symbolically and applies the dual-norm bound in Equation~\eqref{eq:robust-condition}. The eight-anchor monolithic regression retains all symbolic intermediate variables and branch constraints; equality with the folded result checks the contraction implementation.

The measured total is 66.61 seconds. The separately timed components round independently to 16.04 seconds for one-time exact contraction and 50.58 seconds for 1,280 folds; their displayed two-decimal values therefore sum to 66.62 seconds. The ground property comparisons take no measurable solver time. This does not mean the verification problem is approximate; it means construction has simplified it to exact arithmetic without search.

\section{Surrogate-Mediated Verification Details}
\label{app:scd}

The surrogate is a small ReLU network trained to reproduce a finite teacher relation. Candidate programs contain counters, toggles, token predicates, threshold detectors, guarded updates, and small state machines. Synthesis is complete only within the selected template family. The quote counterexample therefore refutes that family, not every possible symbolic program.

The external surrogate and installed program head should not be conflated. The former describes a fixed circuit from outside. The latter is the artifact's actual inference-time attention mechanism. Both can be checked with the same proof-or-counterexample interface, but only the installed program supports the constrained-calibration identity.

\section{Attention as Verifiable Aggregation}
\label{app:deepsets}

Attention can be viewed as query-conditioned aggregation over a multiset of key-value pairs. Let $x_i=(q,k_i,v_i)$ with $q$ fixed across elements. Under suitable assumptions, a permutation-invariant operator has a Deep Sets form \cite{zaheer2017deepsets}:
\begin{equation}
    A(q,K,V)=\rho\!\left(\sum_i \phi(q,k_i,v_i)\right).
\end{equation}
A convenient, but not theorem-forced, factorization is
\begin{equation}
    \phi(q,k_i,v_i)=u(q,k_i)\,v(k_i,v_i),
\end{equation}
where $u$ is a scalar relevance function and $v$ a content map. Restricting $u$, $v$, and $\rho$ to affine maps, piecewise-linear functions, Boolean selection, and tractable aggregation defines a broader search space of verifiable attention operators. Program heads instantiate an extreme point in this space: relevance is a discrete symbolic rule rather than a neural QK product.

\section{Reproducibility and Evidence Release}
\label{app:repro}

The public code and evidence release is:
\begin{itemize}[leftmargin=*]
    \item \href{https://github.com/neelsomani/verifiable-transformers}{\nolinkurl{github.com/neelsomani/verifiable-transformers}}
    \item \href{https://github.com/neelsomani/verifiable-transformers/releases/tag/v1.0}{\nolinkurl{github.com/neelsomani/verifiable-transformers/releases/tag/v1.0}}
    \item \href{https://github.com/neelsomani/symbolic-circuit-distillation}{\nolinkurl{github.com/neelsomani/symbolic-circuit-distillation}}
\end{itemize}

Release \texttt{v1.0} contains the protocols, compact artifacts, evidence manifest, and per-input verification records underlying the experiments reported in this paper. The GPT-2 calibrated checkpoint is archived externally and pinned by SHA-256 in \path{artifacts/gpt2-phase-q-agent/evidence_manifest.json}. The manifest identifies the program-local $W_V/W_O$ calibration rung as the model of record and records the hashes of every frozen non-program parameter. The release documentation \texttt{docs/SCALABILITY.md} and \texttt{docs/VERIFIED\_DISTILLATION.md} give commands and artifact provenance.

\section*{Acknowledgments}
This manuscript was drafted with assistance from OpenAI GPT-5.6 Pro. All technical content, experiments, and claims are the author's.

\end{document}